\documentclass[letterpaper, 10pt, conference]{ieeeconf}      

\IEEEoverridecommandlockouts                              

\usepackage{etex} 
\usepackage{tikz}
\usetikzlibrary{decorations.markings}
\usetikzlibrary{%
    decorations.pathreplacing,%
    decorations.pathmorphing%
}
\usepackage{gensymb}

\usepackage{caption}
\usepackage{array}
\newcolumntype{P}[1]{>{\centering\arraybackslash}p{#1}}
\newcolumntype{M}[1]{>{\centering\arraybackslash}m{#1}}

\usepackage{mathtools}

\usepackage{array}
\newcolumntype{P}[1]{>{\centering\arraybackslash}p{#1}}
\newcolumntype{M}[1]{>{\centering\arraybackslash}m{#1}}

\usepackage{lipsum}
\usepackage[utf8]{inputenc}
\usepackage{makeidx}
\usepackage{amsmath,bm}
\usepackage{siunitx}
\usepackage{amssymb}
\usepackage{color, colortbl}
\usepackage{ifpdf}
\usepackage{wrapfig}
\usepackage{graphicx}
\usepackage{sidecap}
\usepackage[]{threeparttable}
\usepackage{multirow}
\usepackage{multicol}
\usepackage{relsize}
\usepackage{algorithm}
\usepackage{algpseudocode}
\usepackage{pgfplots}
\usepackage{mathrsfs}
\usepackage{bbm}
\usepackage{mathtools}
\usepackage{upgreek}
\usepackage{trfsigns}
\usepackage{arcs}
\usepackage{pgf}
\usepackage{tikz}
\usepackage{lscape}
\usepackage{array, makecell} %
\usepackage{todonotes}
\usepackage{caption}
\usepackage{subcaption}
\usepackage{svg}
\usepackage{cite}
\usepackage{soul}
\usepackage{comment}
\usepackage{import} 
\usepackage{amsmath}
\usepackage{amssymb}
\DeclareMathOperator{\diag}{diag}
\usepackage[section]{placeins} 
\usepackage{float}             
\usepackage{balance}
\usepackage{stfloats}

\usetikzlibrary{shapes.geometric,arrows,automata,positioning}
\usepackage[figuresright]{rotating}
\usepackage[stable, hang]{footmisc}
\usepackage{xparse,mathtools}
\usepackage{xstring}
\ifpdf
  \DeclareGraphicsExtensions{.pdf,.jpg,.png,.ps,.eps}
\else
  \DeclareGraphicsExtensions{.eps,.ps,.png,.jpg,.pdf}
\fi
\usepackage[normalem]{ulem}
\DeclareMathAlphabet{\mathbfsf}{\encodingdefault}{\sfdefault}{bx}{sl}
\DeclareMathAlphabet{\mathitsf}{\encodingdefault}{\sfdefault}{m}{sl}

\ifx\isieee 

	\renewcommand{\eqref}[1]{(\ref{#1})}

\else

\fi

\tikzstyle{startstop} = [rectangle, rounded corners, minimum width=3cm, minimum height=1cm,text centered, draw=black, thick, fill=red!30]
\tikzstyle{io} = [trapezium, trapezium left angle=70, trapezium right angle=110, minimum width=3cm, minimum height=1cm, text centered, draw=black, thick, fill=blue!30]
\tikzstyle{process} = [rectangle, minimum width=3cm, minimum height=1cm, text centered, draw=black, thick, fill=orange!30]
\tikzstyle{decision} = [diamond, aspect=3, minimum width=3cm, minimum height=1cm, text centered, draw=black, thick, fill=green!30]
\tikzstyle{arrow} = [thick,->,>=stealth]
\tikzstyle{noarrow} = [thick,-]
\tikzstyle{doublearrow} = [thick, double distance=1pt, ->, >=stealth]
\tikzstyle{doublenoarrow} = [thick, double distance=1pt]
\tikzstyle{innerwhite} = [thin, white, line width=1pt]
\tikzset{state/.style={
           rectangle,
           rounded corners,
           draw=black, very thick,
           minimum height=2em,
           inner sep=2pt,
           },}
\tikzstyle{start} = [circle, inner sep=0.06cm, draw=black, fill=black]      

\tikzstyle{gain} = [draw, very thick, fill=white, regular polygon, regular polygon sides=3, shape border rotate=30, minimum height=3em, minimum width=3em]
\tikzstyle{bigblock} = [draw, thick, fill=white, rectangle, 
minimum height=4em, minimum width=6em]
\tikzstyle{smallblock} = [draw, thick, fill=white, rectangle, 
minimum height=1em, minimum width=1em]
\tikzstyle{medianblock} = [draw, thick, fill=white, rectangle, 
minimum height=3em, minimum width=3em]
\tikzstyle{bigfcnblock} = [draw, thick, fill=white, rectangle, 
minimum height=4em, minimum width=6em, double, double distance=0.5mm]
\tikzstyle{smallfcnblock} = [draw, thick, fill=white, rectangle, 
minimum height=1em, minimum width=1em, double, double distance=0.5mm]
\tikzstyle{medianfcnblock} = [draw, thick, fill=white, rectangle, 
minimum height=3em, minimum width=3em, double, double distance=0.5mm]
\tikzstyle{invisiblenode} = [draw=none, minimum size=0pt]
\tikzstyle{sum} = [draw, thick, fill=white, circle, node distance=1cm]
\makeatletter
\pgfdeclareshape{record}{
	\inheritsavedanchors[from={rectangle}]
	\inheritbackgroundpath[from={rectangle}]
	\inheritanchorborder[from={rectangle}]
	\foreach \x in {center,north east,north west,north,south,south east,south west}{
		\inheritanchor[from={rectangle}]{\x}
	}
	\foregroundpath{
		\pgfpointdiff{\northeast}{\southwest}
		\pgf@xa=\pgf@x \pgf@ya=\pgf@y
		\northeast
		\pgfpathmoveto{\pgfpoint{0}{0.33\pgf@ya}}
		\pgfpathlineto{\pgfpoint{0}{-0.33\pgf@ya}}
		\pgfpathmoveto{\pgfpoint{0.33\pgf@xa}{0}}
		\pgfpathlineto{\pgfpoint{-0.33\pgf@xa}{0}}
		\pgfpathmoveto{\pgfpointadd{\southwest}{\pgfpoint{-0.33\pgf@xa}{-0.6\pgf@ya}}}
		\pgfpathlineto{\pgfpointadd{\southwest}{\pgfpoint{-0.5\pgf@xa}{-0.6\pgf@ya}}}
		\pgfpathlineto{\pgfpointadd{\northeast}{\pgfpoint{-0.5\pgf@xa}{-0.6\pgf@ya}}}
		\pgfpathlineto{\pgfpointadd{\northeast}{\pgfpoint{-0.33\pgf@xa}{-0.6\pgf@ya}}}
	}
}
\makeatother
\tikzstyle{saturation} = [draw, record, fill=white, minimum height=3em, minimum width=3em]

\tikzstyle{spring}=[thick,decorate,decoration={zigzag,amplitude=0.2cm,pre length=0.3cm,post length=0.3cm,segment length=6}]
\tikzstyle{damper}=[thick,decoration={markings,  
  mark connection node=dmp,
  mark=at position 0.5 with 
  {
    \node (dmp) [thick,inner sep=0pt,transform shape,rotate=-90,minimum width=15pt,minimum height=3pt,draw=none] {};
    \draw [thick] ($(dmp.north east)+(2pt,0)$) -- (dmp.south east) -- (dmp.south west) -- ($(dmp.north west)+(2pt,0)$);
    \draw [thick] ($(dmp.north)+(0,-5pt)$) -- ($(dmp.north)+(0,5pt)$);
  }
}, decorate]
\tikzstyle{ground}=[fill,pattern=north east lines,draw=none,minimum width=0.75cm,minimum height=0.3cm]

\tikzset{
	pics/.cd,
	vector out/.style={
		code={
			\draw[#1] (0,0)  circle (1) (45:1) -- (225:1) (135:1) -- (315:1);
		}
	}
}
\tikzset{
	pics/.cd,
	vector in/.style={
		code={
			\draw[#1] (0,0)  circle (1);
			\fill[#1] (0,0)  circle (.1);
		}
	}
}

\tikzstyle{circlenode} = [draw, thick, circle, minimum width=10pt]

\newcommand{\djallil}[1]{{\color{magenta}{}{Djallil: #1}}}
\newcommand{\jon}[1]{{\color{blue!70}{}{Jon: #1}}}

\renewcommand{\baselinestretch}{0.95}

\usepackage[compact]{titlesec}

\titlespacing{\section}{0pt}{1.0ex plus .5ex minus .2ex}{0.7ex plus .5ex minus .2ex}
\titlespacing{\subsection}{0pt}{1.0ex plus .5ex minus .2ex}{0.7ex plus .5ex minus .2ex}
\titlespacing{\subsubsection}{0pt}{1.0ex plus .5ex minus .2ex}{0.7ex plus .5ex minus .2ex}

\makeatletter
\let\NAT@parse\undefined
\makeatother
\usepackage{hyperref}
\ifpdf
  \usepackage{hyperref}
\else
  \usepackage[hypertex]{hyperref}
\fi

\definecolor{dgreen}{rgb}{0,0.7,0}
\definecolor{dblue}{rgb}{0,0,0.60}
\definecolor{dred}{rgb}{0.7,0,0}
\hypersetup{bookmarksnumbered=true,
            plainpages=false,
            colorlinks=true,
            linkcolor=dblue,
            citecolor=dblue,
            filecolor=dblue,
            urlcolor=dblue}

\title{\LARGE \bf
Bridging the Sim-to-Real Gap with multipanda\_ros2: \\A Real-Time ROS 2 Framework for Multimanual Systems
}
\author{Jon \v{S}kerlj\textsuperscript{1$\dagger$}, Seongjin Bien\textsuperscript{2}, Abdeldjallil Naceri\textsuperscript{1} and Sami Haddadin\textsuperscript{3}
\thanks{© 2026 IEEE. Personal use of this material is permitted. Permission from IEEE must be obtained for all other uses, including reprinting/republishing this material for advertising or promotional purposes, collecting new collective works for resale or redistribution to servers or lists, or reuse of any copyrighted component of this work in other works.}
\thanks{
This work was supported by the Lighthouse Initiative Geriatronics by StMWi Bayern (Project X, grant no. IUK-1807-0007// IUK582/001) and LongLeif GaPa gGmbH (Project Y).}
\thanks{$^{1}$Munich Institute of Robotics and Machine Intelligence (MIRMI), Technische Universität München (TUM), Germany.}
\thanks{$^{2}$ Technische Universität Nürnberg (UTN), Germany}
\thanks{$^{3}$ Mohamed Bin Zayed University of Artificial Intelligence, UAE}
\thanks{
\textsuperscript{$\dagger$}Corresponding author. Email: \texttt{jon.skerlj@tum.de}}
}

\begin{document}

\maketitle
\thispagestyle{empty}
\pagestyle{empty}


\begin{abstract}


We present \texttt{\href{ https://github.com/tenfoldpaper/multipanda_ros2}{multipanda\_ros2}}, a novel open-source ROS 2 architecture for multi-robot control of Franka Robotics robots. Leveraging \texttt{ros2\_control}, this framework provides native ROS 2 interfaces for controlling any number of robots from a single process. Our core contributions address key challenges in real-time torque control, including interaction control and robot-environment modeling. A central focus of this work is sustaining a 1kHz control frequency, a necessity for real-time control and a minimum frequency required by safety standards. Moreover, we introduce a controllet-feature design pattern that enables controller-switching delays of $\le 2$ ms, facilitating reproducible benchmarking and complex multi-robot interaction scenarios. To bridge the simulation-to-reality (sim2real) gap, we integrate a high-fidelity MuJoCo simulation with quantitative metrics for both kinematic accuracy and dynamic consistency (torques, forces, and control errors). Furthermore, we demonstrate that real-world inertial parameter identification can significantly improve force and torque accuracy, providing a methodology for iterative physics refinement. Our work extends approaches from soft robotics to rigid dual-arm, contact-rich tasks, showcasing a promising method to reduce the sim2real gap and providing a robust, reproducible platform for advanced robotics research.
\end{abstract}
\vspace{-2mm}
\begin{flushleft}
\noindent\textbf{Published version:} 2026 IEEE International Conference on Robotics and Automation (ICRA),\\
DOI: \texttt{\href{https://doi.org/10.1109/ICRA57385.2026.11696149}{10.1109/ICRA57385.2026.11696149}}
\end{flushleft}
\section{Introduction}
For robots to be able to operate in the challenging, dynamic and cluttered environments they are used in today, there are high requirements towards their control, among them: Torque control, Real-time capability, and the ability to quickly change between controllers to adapt to new tasks, constraints and environments.  
The foundation for robots to sense was laid with force and torque control~\cite{Whitney1977}, which represents a critical facet of robotics and automation, serving as a key enabler for robots to interact with their environments in a refined and intelligent manner. 
The capability to precisely control force and torque is indispensable for performing delicate tasks, ranging from assembly operations in industrial settings~\cite{Gasparetto2019} to surgical procedures~\cite{George2018}. 

The early important work on giving robots a sense of touch and variable stiffness was performed by Salisbury~\cite{Salisbury1980} in 1980, which allowed the control of the stiffness of and all 6 degrees of freedom with a control rate of 60~Hz. 
In further research~\cite{Craig1981}, this approach was extended to allow for a separation of directions where either force or position could be controlled. 
In these experiments, an improved force response could be observed by increasing the servo rate from 60~Hz to 120~Hz, but it was still unable to control the higher oscillations at about 280~Hz, due to computational limitations. 
With the development of more lightweight and flexible robots, the frequency of these oscillations further increased~\cite{Troch1988} leading to a necessity of control frequency of up to 1~kHz~\cite{Hirzinger2002},\cite{Haddadin2022}.

Controlling the robots for safe and effective interaction now goes beyond traditional positional and force-based control, e.g. impedance control~\cite{Hogan1984} in scenarios where the robot requires compliance, or the converse, admittance control~\cite{Newman1992}, mapping force inputs to motion outputs. 
Even more complex tasks like wiping a surface require unified frameworks~\cite{AlbuSchaeffer2007} for successful task execution. 
Other scenarios like hand-over or the manipulation of unwieldy objects involve two or more robots in synchronous, cooperative, or coupled movement~\cite{Slotine2009}, \cite{Laha2024}. Naturally, scenarios in which a combination of such skills is required would inevitably call for the robot to be capable of rapidly and seamlessly switching from one skill to another.
\begin{figure}[t]%
    \vspace{4mm}
	\centering
    \subfloat{{\includegraphics[width=0.99\columnwidth]{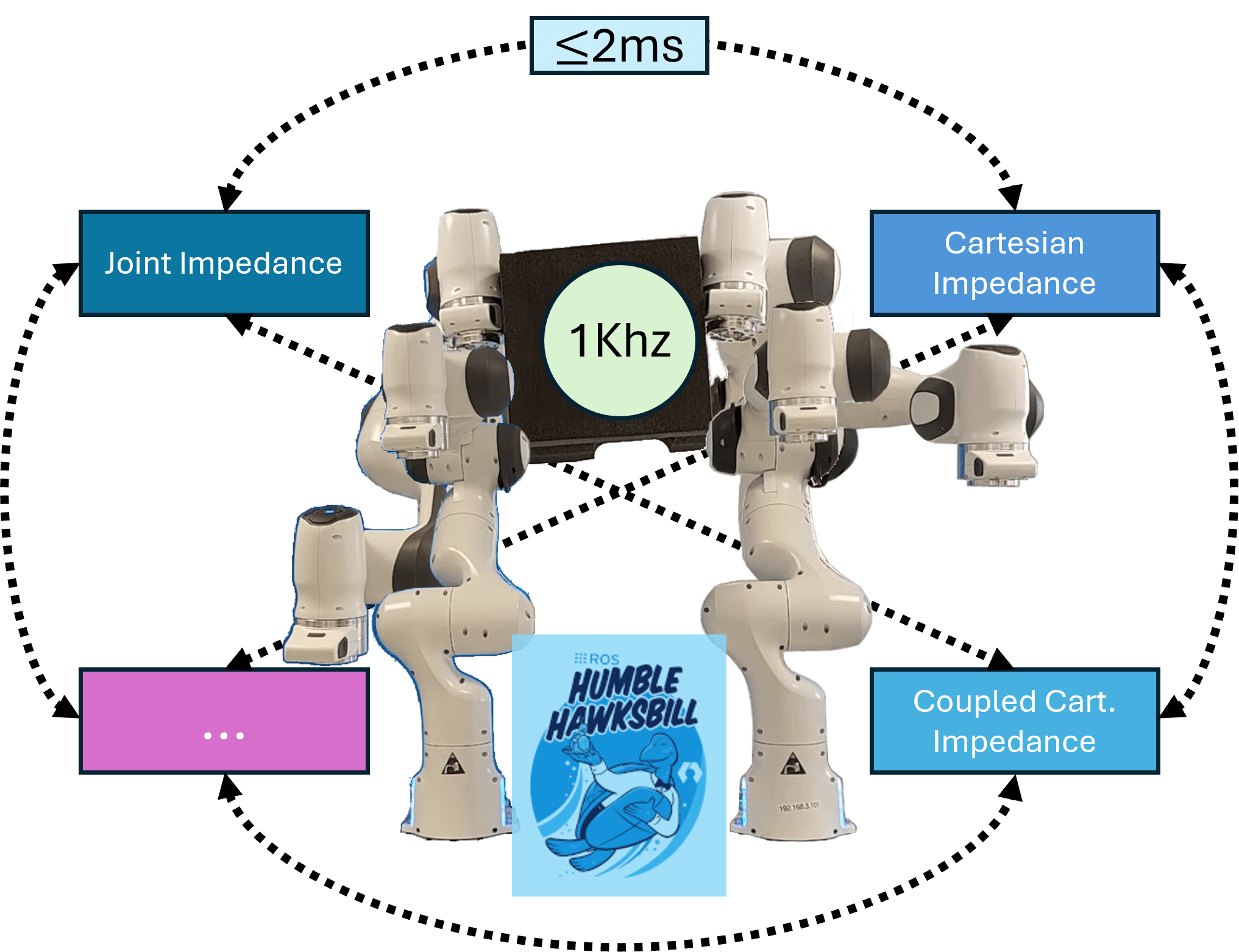}}}
	\caption{An overview of \texttt{multipanda\_ros2}, summarizing its core features.
    }
	\label{fig:overview_multipanda}
\end{figure}

Having discussed the different control types, the importance of maintaining a high control sample rate, and the necessity for ensuring flexible controller interchangeability, outlined in Fig.~\ref{fig:overview_multipanda}, the novel contributions of our work can be stated as follows:
\begin{itemize}
    \item Review real-time torque control for single- and dual-arm robots, emphasizing interaction control and robot–environment modeling challenges.
    \item Introduce an open-source ROS 2 multirobot framework for Franka Robotics robots with $\leq 2$ ms controllet-switching delays for reproducible benchmarking.
    \item Integrate MuJoCo simulations with metrics for kinematic accuracy and dynamic consistency (torques, forces, control errors).
    \item Assess force-based fidelity, extending soft-robotics approaches \cite{Gao2024ResidualPhysics} to rigid dual-arm, contact-rich tasks.
    \item Real-world inertial parameter identification reduces the sim-to-real gap, improving force/torque accuracy and enabling iterative physics refinement.
\end{itemize}

\section{State of the Art}
\begin{table*}[!b]
\begin{center}
\begin{tabular}{*{15}{c}}
    \\
    \hline
    Name & Structure & Year & Command $\tau_{d}$ freq. & Internal freq. & Sensor freq. & Middleware \\
    \hline
    DLR LWR I \cite{AlbuSchaeffer2000}& Single & 2000 & 1~kHz & 3~kHz & 1~kHz & Custom \\
    DLR LWR II \cite{Hirzinger2001}& Single & 2001 & 1~kHz & 3~kHz & 1~kHz & Custom \\
    DLR LWR III \cite{Hirzinger2002}& Single & 2002 & 1~kHz & 3~kHz & 1~kHz & Custom \\
    DLR Justin \cite{Ott2006}& Humanoid & 2006 & 1~kHz & 3~kHz & 1~kHz & Custom \\
    DLR-Kuka LWR \cite{Bischoff2010}& Single & 2008 & 1~kHz & 3~kHz & 1~kHz & Sunrise.OS \\
    Rethink Robotics Baxter \cite{RethinkRobotics2022ArmControl}& Bimanual & 2011 & 1~kHz & - & 1~kHz & ros\_control \\
    Barrett WAM 7DoF \cite{Barret2011} & Single & 2011 & 1~kHz & - & $\geq$1~kHz & Custom \\
    Boston Dynamics Atlas 1 \cite{Feng2014}& Humanoid & 2013 & 333~Hz & 1~kHz & - & Custom \\
    NASA Valkyrie \cite{Radford2015} & Humanoid & 2013 & 1~kHz & 5~kHz & 1~kHz & ros\_control \\
    DLR TORO \cite{Henze2014}& Humanoid & 2014 & 1~kHz & 3~kHz & 1~kHz & Custom \\
    Rethink Robotics Sawyer \cite{RethinkRobotics2022ArmControl}& Single & 2015 & 1~kHz & - & 1~kHz & ros\_control \\
    PAL Robotics TIAGo \cite{Pages2016}& Humanoid & 2016 & 1~kHz & - & 1~kHz & ros(2)\_control \\
    PAL Robotics TALOS \cite{Stasse2017}& Humanoid & 2017 & 1~kHz & - & 1~kHz & ros\_control \\
    KIT ARMAR-6 \cite{Asfour2018}& Humanoid & 2018 & 1~kHz & - & 1~kHz & Custom \\
    Doosan M-Series \cite{DoosanRobotics2024M0609}&  Single & 2018 & - & - & $\geq$1~kHz & Custom \\
    Berkeley Blue \cite{Gealy2019}& Bimanual & 2019 & 170~Hz & - & 20~kHz (Current) & ros\_control \\
    Franka Robotics robot \cite{Haddadin2022}& Single & 2019 & 1~kHz & 3~kHz & 1~kHz & ROS, Custom \\
    DLR SARA \cite{Iskandar2020}& Single & 2021 & - & 8~kHz & - & Custom \\
    GARMI \cite{Trobinger2021}& Humanoid & 2021 & 1~kHz & 3~kHz & 1~kHz & ros\_control\\
    KINOVA Gen3 \cite{Kinova2022} & Single & 2022 & 1~kHz & - & $\geq$1~kHz & Custom \\
    1X Eve \cite{1XTechnologiesEve2023}& Humanoid  & 2023 & - & - & - & Custom \\
    Han's Robots Elfin Pro Series \cite{ElfinPro2023} & Single & 2023 & 1~kHz & - & $\geq$1~kHz & Custom \\
    Apptronik Apollo \cite{Apollo2024} & Humanoid & 2024 & - & - & - & Custom \\
    \hline
\end{tabular}
\caption{A non-exhaustive list of torque-controlled commercial and research robots. To be included, a robot must explicitly state that it supports torque or effort command. For commercial robots, only those certified as cobots are considered. Unidentified values are marked with '-'.}
\label{Table:robots}
\end{center}
\end{table*}
\subsection{High Frequency Torque Control} 
High-frequency torque control is fundamental for safe and accurate physical human--robot interaction. A 1~kHz low-level torque command loop enables robots to be modeled according to Lagrangian dynamics, such that higher-level controllers can treat the joints as ideal torque sources \cite{Haddadin2022}. At the same time, safety standards such as ISO 10218 \cite{ISO10218} specify 1~kHz measurement frequencies for collision detection to minimize injury risk, which has driven widespread adoption of this frequency in collaborative robots (cobots). Prominent examples include the DLR Lightweight Robot series \cite{AlbuSchaeffer2000,AlbuSchaeffer2007}, Franka Emika Panda \cite{Haddadin2022}, KINOVA Gen 3 \cite{Kinova2022}, and Barrett WAM \cite{Barret2011}.  

These systems often employ cascaded architectures \cite{Iskandar2020}, where an outer-loop controller generates task-level commands and inner loops regulate motor torques or velocities. This design improves precision and disturbance rejection while retaining fast safety responses. At the joint level, torque control is critical to guarantee safe human interaction \cite{Haddadin2007}. However, as noted by \cite{Kirschner2021}, torque sensing alone is insufficient for fine-grained force control in contact-rich environments, motivating integrated strategies that combine sensing, impedance regulation, and high-rate control.





\subsection{Existing Multimanual Architectures}
Dual-arm and humanoid platforms impose stricter real-time requirements due to higher degrees of freedom and the need for coordinated control. Middleware frameworks such as ROS and ROS 2 \cite{Koubaa2017,ros2}, extended with packages like \texttt{ros\_control} \cite{ros_control}, have become de facto standards for modular development, while OROCOS \cite{Bruyninckx2001} and custom-built architectures provide low-latency real-time support.  

ROS is widely used across many robot configurations, including dual-arm systems such as Baxter \cite{RethinkRobotics2022ArmControl}, GARMI \cite{Trobinger2021}, and Berkeley Blue \cite{Gealy2019}. The \texttt{ros\_control} package \cite{ros_control} enables real-time, hardware-agnostic control, allowing the same code to run seamlessly on different robots. ROS 2~\cite{ros2} improves on ROS by enhancing real-time capability, security, and robustness, introducing node composition \cite{ros2_node_composition} and leveraging data distribution service (DDS) \cite{Liu2018} for high-performance data exchange to meet strict timing requirements in modern robotics.



Custom software enables 1~kHz control loops with minimal latency, critical for safety and for bridging the sim2real gap. DLR’s bimanual Justin \cite{Ott2006} uses a component-based Robot Development concept \cite{Bauml2006} with modular subsystems and a dedicated real-time system for precise control. Similarly, KIT’s ARMAR-6 \cite{Asfour2018} employs ArmarX \cite{Vahrenkamp2015} for distributed parallel processing and tightly integrated real-time control. Table~\ref{Table:robots} summarizes humanoid setups, control frequencies, and software frameworks.

\subsection{Robotic simulators}
Simulation plays a crucial role in modern robotics, enabling researchers and engineers to design, test, and validate algorithms in a safe, cost-effective environment. It allows rapid prototyping without the risk of damaging hardware and accelerates development by supporting parallel testing of control strategies, perception pipelines, and interaction scenarios. Moreover, simulation makes it possible to explore edge cases and rare events that would be expensive or unsafe to reproduce on physical robots.

The field of robotics currently supports a wide range of simulators. A few of the most known are Gazebo \cite{2004Koenig}, MuJoCo \cite{todorov2012mujoco}, Webots \cite{1998Webots}, CoppeliaSim \cite{2013RohmerCoppelia}, NVIDIA Isaac Sim \cite{NVIDIA_Isaac_Sim}, PyBullet \cite{coumans2016pybullet}, and Pinocchio \cite{pinocchio2019}, each designed with specific requirements in mind and varying in fidelity, ease of integration, and computational efficiency \cite{2021Collins,FARLEY2022mobileSimulator}. MuJoCo and Pinocchio are often preferred for high-frequency, physics-accurate simulations: MuJoCo for control-intensive tasks and sim-to-real benchmarking, and Pinocchio for fast rigid-body dynamics computations, particularly in multi-body or humanoid systems, as summarized in Table~\ref{Table:robots}.

The sim-to-real gap is especially prevalent in the field of reinforcement learning, where policies trained in simulation often fail on real systems. Current work mainly tackles this by making policies robust to discrepancies, for example, through transfer learning and domain adaptation, without improving the simulator itself \cite{2020ZhaoRL}. To the best of our knowledge, the examples of closing the gap from the simulation side, using real-world data to iteratively update and refine simulators, are scarce. One exception is the work of \cite{Gao2024ResidualPhysics} which utilizes residual physics networks to learn from real data and reduce the sim-to-real gap. However, additional research is needed to see the full effect of this approach on the sim-to-real gap challenge.

\section{Implementation}
\begin{figure*}[t]%
	\centering
    \includegraphics[width=0.9\linewidth]{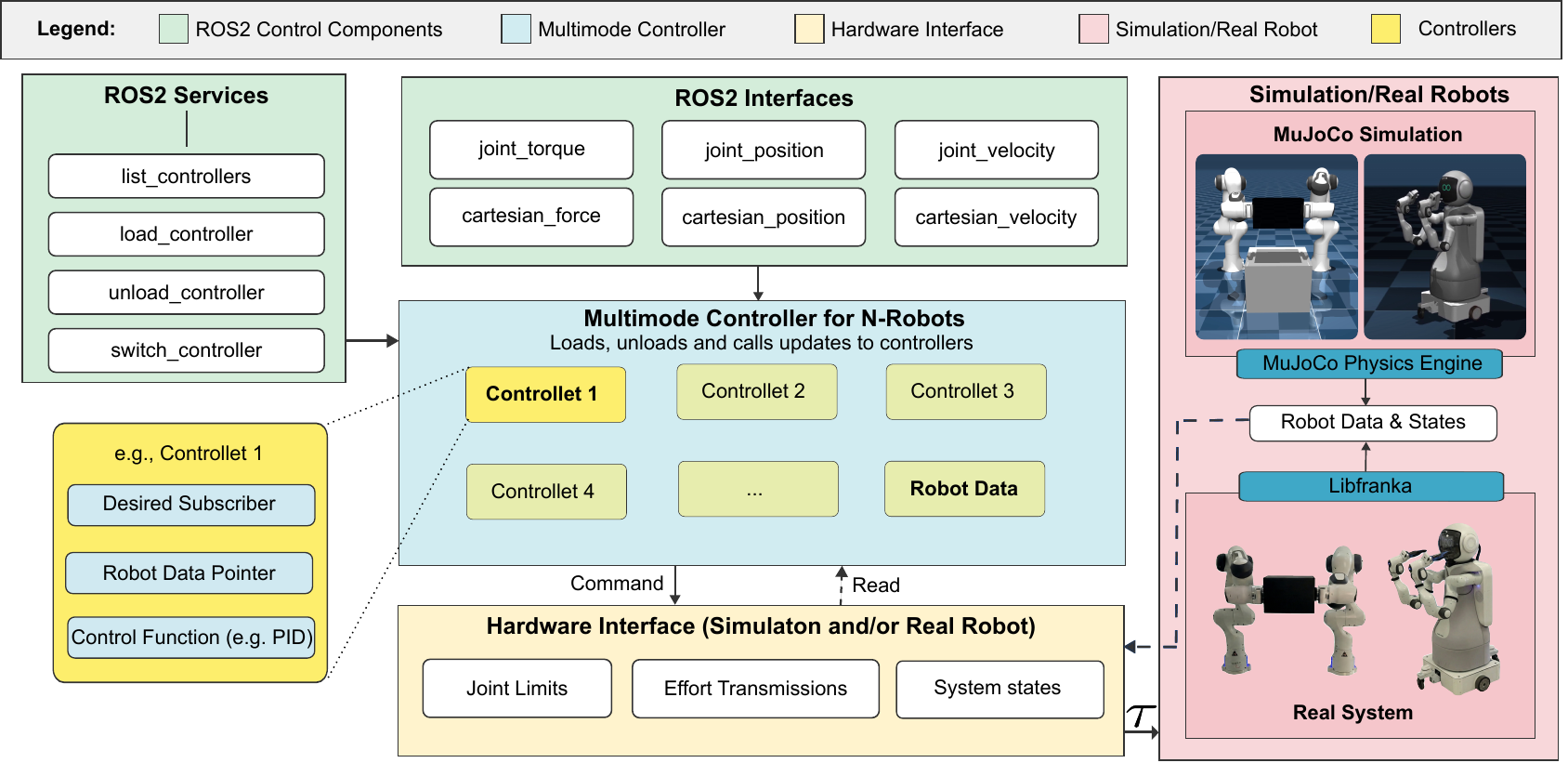}
	\caption{Architecture diagram of the \texttt{multipanda\_ros2} within \texttt{ros2\_control} framework.}
	\label{fig:architecture}
\end{figure*}
In this section, we present \texttt{\href{ https://github.com/tenfoldpaper/multipanda_ros2}{multipanda\_ros2}}\footnote{\url{https://github.com/tenfoldpaper/multipanda\_ros2}}, a software framework developed for the coordinated control of multiple Franka Robotics robots. Our framework is built upon the \texttt{ros2\_control} infrastructure, as illustrated in Fig.~\ref{fig:architecture}, and is fully compatible with the ROS~2 Humble distribution.

The primary goal of \texttt{multipanda\_ros2} is to provide the existing community of Franka Robotics robot users with easy access to ROS 2, while also enabling the full list of features that are provided through the robot's Franka Control Interface (FCI) and \texttt{libfranka}, and to ease the entry into doing bimanual or multimanual torque-control research. The open-source repository includes extensive tutorials and examples to further aid this effort.

The \texttt{multimode\_controller} (MC), shown in the blue block of Fig.~\ref{fig:architecture}, plays a central role in the proposed framework. It is implemented as a typical \texttt{ros2\_control} controller, meaning it can be used alongside other controllers without issue. The MC itself does not implement a controller; instead, it uses \textit{controllets}. Each controllet takes ownership of one or more robots in the \texttt{ros2\_control} framework specified at runtime, and implements the actual command torque $\bm{\tau}_{\text{cmd}}$ calculation for the loop. The MC can be instantiated with an arbitrary number of such controllets, and handles the robot assignment for all the controllets, ensuring that only one controllet has ownership of a robot at any given time. 

The proposed framework has been designed with minimal overhead in mind, making heavy use of move operators and references to avoid unnecessary copying. The MC is instantiated with a vector of controllets upon initialization based on the user's configuration. Each element contains information about the robot that it needs to control, and references to the shared $\bm{\tau}_{\text{cmd}}$ vector and robot data that are centrally managed by the MC. Such a design allows a controllet to have real-time access to the controlled robots' data, which is crucial for algorithms that require the full state and dynamic information of the involved robots. Additionally, this ensures that there are minimal copy operations in the framework, so that the $\bm{\tau}_{\text{cmd}}$ calculation across controllets and the final commanded torque in the robot's \texttt{libfranka} loop stays consistent.

The 1~kHz control loop inside the MC calls the control loop function of the currently active controllets for each step. This enables the implementation of a switching function, which allows the MC to change the currently active controllet at runtime, with a delay of about 2ms (see section \ref{sec:controller_perf} for more details). The controllet switching function is exposed as a ROS 2 service, and the MC ensures that the new set of activated controllers does not have conflicting resource requirements. While it is more desirable to create a single controller that is capable of handling all the tasks that a robot is expected to perform, custom controllers are written for different tasks in practice. As such, this feature allows for complex behaviors in scenarios that require rapid transitions between different controllers to be implemented effectively.

One additional benefit of the proposed structure is the significantly reduced amount of the required boilerplate coding. The controllets make heavy use of C++ templating and inheritance to achieve this. Typically, a communication-less instance of the controllet is implemented, which performs the desired $\bm{\tau}_{\text{cmd}}$ calculation for a given desired pose, and defines the parameter update function. Then, this class is inherited by a ROS 2-enabled instance, which implements the callback functions for desired pose subscription and parameter handling. This allows control algorithms to be only implemented once, then inherited to create a wide range of custom ROS 2 instances whose subscriptions match the particular task. Finally, these controllets can then be loaded into the MC with a simple \texttt{yaml} file update.

Given that \texttt{ros2\_control} abstracts away the robot's interfaces, the proposed MC controller is, in fact, agnostic to the model of the robots it is controlling. This means that the proposed framework can be easily extended to other robots that already have their own \texttt{ros2\_control} frameworks, such as the Baxter~\cite{RethinkRobotics2022ArmControl} and the TIAGo~\cite{Pages2016}. Additionally, while the framework is currently only implemented for the torque control mode of the Franka Robotic robot, extensions to the other control modes are planned.

\subsection{Simulation Environment}
Our primary research focus is physical human–robot interaction (pHRI), which involves high-frequency, control-intensive tasks. Accordingly, we selected the MuJoCo simulator \cite{todorov2012mujoco} due to its efficient handling of dynamics and contact interactions at rates comparable to real hardware, enabling precise torque control and fast feedback loops essential for stable and responsive pHRI experiments.

The simulation is implemented as a plugin to the work of \cite{2025leinsMuJoCoROS}, which wraps the MuJoCo physics engine into a ROS 2 package. This allows development within the same ROS 2 framework used for real hardware, ensuring consistency between simulated and physical experiments. The setup also supports multi-arm simulations, enabling the same controllers to be run on both simulated and real robots. As with the real-hardware framework, control in simulation relies on \texttt{ros2\_control}, facilitating straightforward extension to other robots that implement their own \texttt{ros2\_control} interfaces. Following these design principles, a controller needs to be developed only once and can run on both simulation and real hardware without any modifications.

\section{Methods}\label{sec:methods}

To evaluate the multimode controller performance and sim-to-real gap in robotic manipulation, we implemented the Unified Force-Impedance Control (UFIC) framework from \cite{haddadin2024} and a classical Cartesian impedance controller (baseline) for both simulation and real-world experiments using Franka Robotics robots. This comparison quantifies performance differences in force tracking, stability, and energy management, assessing simulation fidelity for real-world deployment.

\subsection{Controller Implementation}
Both controllers were coded in C++ with real-time interfacing via the Franka Emika Fast Research Interface (FRI) at 1 kHz. The baseline impedance controller uses:
\begin{IEEEeqnarray}{rCl}\label{eq:impedance}
\bm{\tau}_i &=& \mathbf{J}^T(\bm{q}) \Bigg[ \mathbf{K}_c \tilde{\bm{x}} + \mathbf{D}_c \dot{\tilde{\bm{x}}} \nonumber  \\
&& + \mathbf{M}_c(\bm{q}) \ddot{\bm{x}}_d + \mathbf{C}_c(\bm{q}, \dot{\bm{q}}) \dot{\bm{x}}_d + \mathbf{F}_g(\bm{q}) \Bigg]. 
\end{IEEEeqnarray}
where $\tilde{\bm{x}} = \bm{x}_d - \bm{x}$, with task torque 
\begin{equation}
    \bm{\tau}_{\text{task}} = \mathbf{J}^T (-\mathbf{K}_c \tilde{\bm{x}} - \mathbf{D}_c (\mathbf{J} \dot{\bm{q}}))
\end{equation} 
and nullspace torque 
\begin{equation}
\bm{\tau}_{\text{null}} = (\mathbf{I} - \mathbf{J}^\dagger \mathbf{J})^T (\mathbf{K}_N (\bm{q}_{d_N} - \bm{q}) - \mathbf{D}_N \dot{\bm{q}})
\end{equation}

The UFIC extends this with force control:
\begin{equation}\label{eq:ufic}
\bm{\tau}_m' = \bm{\tau}_i' + \bm{\tau}_f', \quad \bm{\tau}_f' = \mathbf{J}^T(\bm{q}) \mathbf{F}_f',
\end{equation}
where $\mathbf{F}_f' = (\gamma_f + \alpha_f (1 - \gamma_f)) \mathbf{F}_f$ (PID output modulated by tank signals $\alpha_f$ and $\gamma_f = \exp(-\frac{\|\bm{x} - \bm{x}_c\|^2}{d_{\max}^2}$)), and $\bm{\tau}_i'$ uses modulated $\tilde{\bm{x}}' = \bm{x}_d' - \bm{x}$ with $\bm{x}_d' = \bm{x}_d + \alpha_i (\bm{x}_d - \bm{x})$. Tank energies are initialized as $E_{f,0} = \frac{1}{2} \mathbf{K}_p \|\mathbf{F}_d\|^2$, $E_{i,0} = E_{f,0}$.

Dual-arm coordination applied a $\pm 15$ cm y-axis offset to goal poses. Additional features included GJK-based self-collision avoidance ($\bm{\tau}_{\text{ca}}$ for distances $<$ 0.05 m) and manipulability-based singularity avoidance:
\begin{align}
m_{\text{kin}}(\bm{q}) &= \sqrt{\det(\mathbf{J}(\bm{q}) \mathbf{J}^T(\bm{q}))}, \\
V_{\text{sing}}(\bm{q}) &= 
\begin{cases}
k_m (m_{\text{kin}}(\bm{q}) - m_0)^2 & m_{\text{kin}}(\bm{q}) \leq m_0 \\
0 & \text{otherwise},
\end{cases} \\
\bm{\tau}_{\text{ma}} &= -\frac{\partial V_{\text{sing}}(\bm{q})}{\partial \bm{q}},
\end{align}
with $k_m = 10$, $m_0 = 0.1$. The final command was:
\begin{equation}
\bm{\tau}_{\text{cmd}} = \bm{\tau}_{\text{task}} + \bm{\tau}_{\text{null}} + \bm{\tau}_{\text{cor}} + \bm{\tau}_{\text{ca}} + \bm{\tau}_{\text{ma}},
\end{equation}
where $\bm{\tau}_{\text{cor}} = \mathbf{C}(\bm{q}, \dot{\bm{q}}) \dot{\bm{q}} + \mathbf{g}(\bm{q})$. Parameters were: $\mathbf{K}_p = 200 \mathbf{I}$ N/m, $\mathbf{K}_d = 10 \mathbf{I}$ Ns/m, $\mathbf{K}_i = 50 \mathbf{I}$ NRs/m; $\mathbf{K}_c = \diag(100,100,50,10,10,10)$ N/m; $\mathbf{D}_c$ for critical damping; tank limits $E_{\min} = 0$ J, $E_{\max} = 100$ J; $d_{\max} = 0.05$ m; no friction feedforward ($\mathbf{F}_{\text{FF}} = 0$).
\section{Experimental Procedure}\label{sec:experiments}
\subsection{Controller performance}
To evaluate the performance of the proposed framework, we conducted a series of experiments. For this purpose, we implemented a multimode controller with 2 distinct controllets: a dual Cartesian impedance controller (DC), and a coupled dual Cartesian impedance controller (CDC), as detailed in section \ref{sec:methods}, to command two Franka Robotics robots connected to the same network. To increase the complexity to the control loop calculation, we additionally implemented the self-collision avoidance algorithm from \cite{exp_GJK} and the manipulability potential algorithm from~\cite{exp_Manip}. This creates 5 conditions that are being tested: 1) the DC controllet with no features (NF), 2) DC with self-collision avoidance (CA), 3) DC with manipulability potential (MA), 4) DC with both features (CA-MA), and 5) the CDC controllet with manipulability (C-MA). These acronyms are used for reference in Fig. \ref{fig:exp_data}.

For control cycle time calculation, we calculated the difference between the start and end of the control function, i.e. after the $\bm{\tau}_{\text{cmd}}$ were calculated for both robots.
The $\bm{\tau}_{\text{cmd}}$ delay was measured by recording both the $\bm{\tau}_{\text{cmd}}$ and the time at the moment the values are written to the shared memory between \texttt{ros2\_control} and the robot's \texttt{libfranka} loop, and the same values within the \texttt{libfranka} control loop. The two datasets were then linearly interpolated across the union of the two datasets' timestamps, and then cross correlation was performed to identify the delay. Finally, for the computer resource measurement, the command \texttt{top -bn1} for the control program was executed every 0.3 seconds and its corresponding CPU and memory usage data were recorded. Additionally, we measured the controllet switching delay from 50 trials by measuring the time difference from when the controller receives the switch request to the first control loop of the new controllet. All experiments were conducted on the ASRock Industrial 4x4 BOX 7735U/D NUC mini-PC with 32GB of DDR5-5800~Mhz RAM.

With the above arrangements, the experiment was conducted by running each controllet for 30 seconds. At the same time, desired poses were calculated and published at 1~kHz on a separate node that implements the circular field motion generator from~\cite{Laha2024}. The metrics were measured at each cycle, resulting in 30000 data points each for the first 2 metrics, and 100 data points for the resource metric. 


\subsection{Simulation Validation}
To evaluate the fidelity of the proposed simulation environment, we designed a series of experiments aimed at progressively challenging the kinematic and dynamic models of a serial manipulator. The rationale behind this staged approach is twofold. First, simple tasks allow us to isolate the effects of pure kinematic modeling errors. Second, gradually introducing dynamic phenomena, such as inertial effects and contact interactions, enables a structured evaluation of how the simulator captures increasingly complex physical behaviors.

To rapidly quantify deviations between simulated and real robot behavior, we employ a set of straightforward yet informative metrics. The primary measure is the root mean square error (RMSE) between the recorded simulation data and experimental measurements, selected for its simplicity and interpretability. Errors are evaluated over the following key variables: 
$\bm{q}\text{ (joint positions)}$, $\dot{\bm{q}} \text{ (joint velocities)}$, $\bm{\tau} \text{ (joint torques)}$, $\bm{x}_{\text{EE}} \text{ (end-effector position)}$, $\bm{F}_{\text{EE}} \text{ (end-effector force)}$, and $C_{err}$ (control error) \footnote{Control error: defined as the difference between the commanded and \\achieved joint torque values.}, 
providing a comprehensive view of perfomance of both kinematics and dynamics.

We designed five validation tasks, each introducing an increasing level of kinematic and dynamic complexity. Each experiment was executed for 10 seconds and repeated 5 times, averaging the results across trials.\\ \textbf{Task 1:} Executes the default joint impedance example controller (following equation \ref{eq:impedance}), provided by the manufacturer, controlling the robot in free space.
\\ \textbf{Task 2:} Executes the default Cartesian impedance example controller (following equation \ref{eq:impedance}), provided by the manufacturer, controlling the robot in free space.
\\ \textbf{Task 3:} Executes a modified force example controller, provided by the manufacturer, which uses a variable desired contact force perpendicular to a flat surface, following a sinusoidal curve between 0 and 30 N for one period, instead of using a constant desired force. 
\\ \textbf{Task 4:} Executes a custom unified force impedance controller as formulated in equation \ref{eq:ufic}, which maintains a constant 9.81 N contact force perpendicular to a flat surface (in Z-direction) while following a circular trajectory in X-Y plane with radius of 10 cm.
\\ \textbf{Task 5:} Executes a custom coupled unified force impedance controller by extending the formulation of equation \ref{eq:ufic} for two arms in order to coordinate them in grasping a box from opposite sides. Each manipulator maintains a constant 20 N force perpendicular to the box surface while jointly following a 20 cm vertical trajectory.

Lastly, to illustrate the potential for improving the simulation fidelity using real-world data, we perform an identification procedure on the real system following the steps of \cite{Troebinger2023} to update the simulated robot's inertial parameters before repeating task 4.

\section{Results}\label{sec:results}

\subsection{Controller performance results} \label{sec:controller_perf}
\begin{figure}[h]%
	\centering
    \subfloat[\centering Average $\bm{\tau_{\text{cmd}}}$ delay. \label{fig:avg_delay}]{{\includegraphics[width=0.80\columnwidth]{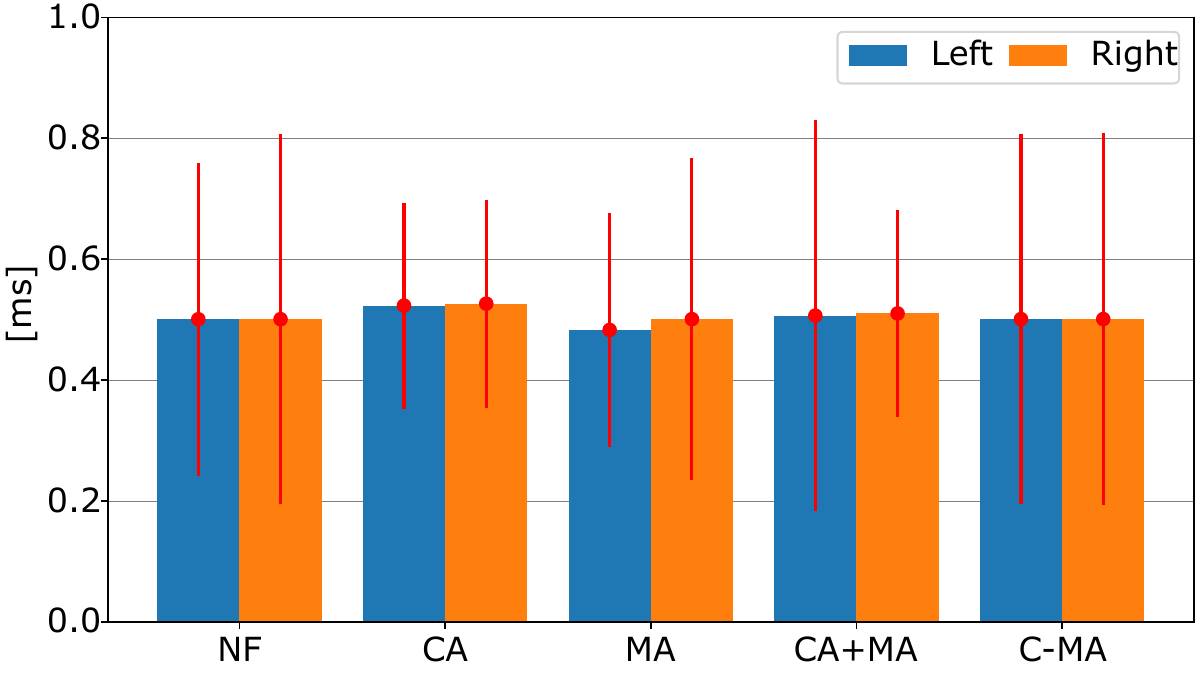} }}%
	\smallskip{}
	\subfloat[\centering Average loop time.\label{fig:avg_loop}]{{\includegraphics[width=0.40\columnwidth]{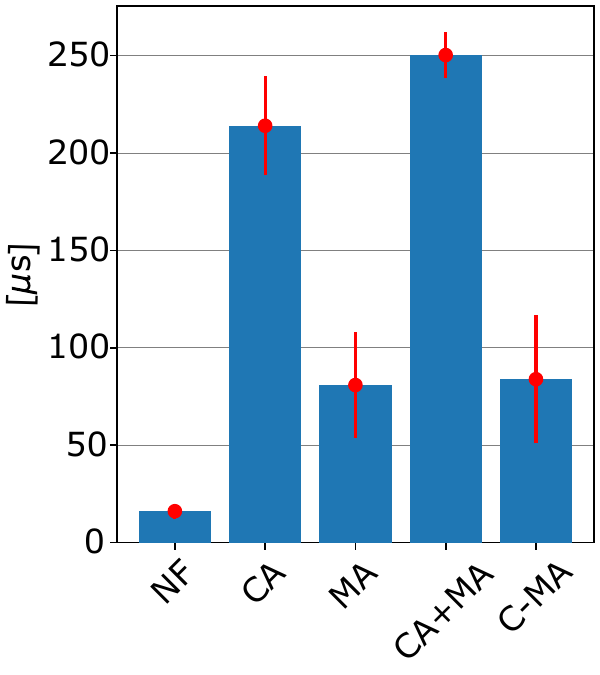} }}
	\smallskip{}
	\subfloat[\centering Average 1-core CPU usage.\label{fig:avg_cpu}]{{\includegraphics[width=0.40\columnwidth]{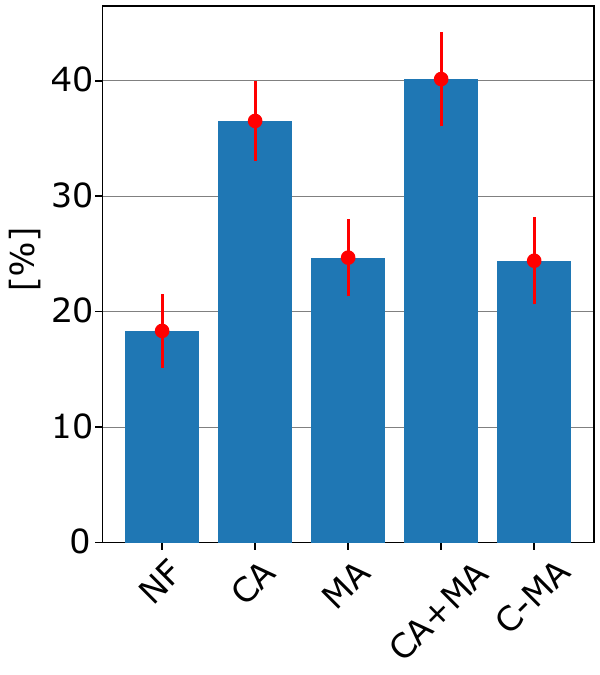} }}%
	\smallskip{}
	\caption{The metrics measured to determine the performance of the proposed framework. The x-axis represents the controllet-feature condition. 
    }
	\label{fig:exp_data}

\end{figure}

The results of the controller performance experiments are summarized in Fig. \ref{fig:exp_data}. We recorded the following metrics: 1) control loop cycle time, 2) time it takes for the computed command torque $\bm{\tau_{\text{cmd}}}$ to reach the robot, and 3) system resource consumption. The above subfigure (\ref{fig:avg_delay}) shows the average $\bm{\tau}_{\text{cmd}}$ delay for each arm which is on average approximately 0.5 ms. This can be explained by the fact that the \texttt{ros2\_control}'s and the \texttt{libfranka}'s control loops run on separate threads. From this, it can be ascertained that most new $\bm{\tau}_{\text{cmd}}$ are consumed immediately by the robot's control loop promptly, with occasional single control cycle delays.

The average loop time is presented in Fig.~\ref{fig:avg_loop}. The  baseline DC controllet requires less than 25$\mu$s per cycle, while the collision avoidance algorithm introduces a noticeable computational overhead, as expected given the presence of two nested \texttt{for} loops (one per arm). In contrast, the manipulability potential calculation adds approximately 40–60~$\mu$s, depending on the condition. Nevertheless, the overall control loop remains sufficiently short to sustain the 1~kHz target frequency, leaving room for additional features before reaching this limit.

The CPU usage, shown in Fig.~\ref{fig:avg_cpu}, corresponds to a single core of the 16-core PC. Across all conditions, memory usage remained constant at 0.2\%, constrained by the single-decimal precision of the \texttt{top} command. Overall, these results indicate that the proposed framework consumes only a modest amount of resources and does not hinder the execution of other tasks.

The controllet switching experiment yielded an average of 2.117ms with a standard deviation of 0.493ms, indicating that it takes about 2 control cycles before the new controllet is activated.

\subsection{Simulation validation results}
\begin{figure}[h]
    \subfloat[Task 1]{\includegraphics[width=.48\columnwidth]{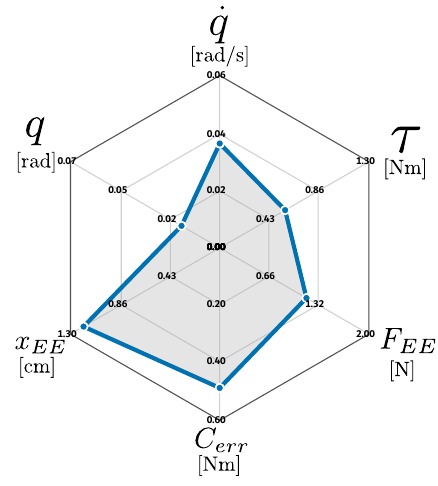}}\hfill
     \subfloat[Task 2]{\includegraphics[width=.48\columnwidth]{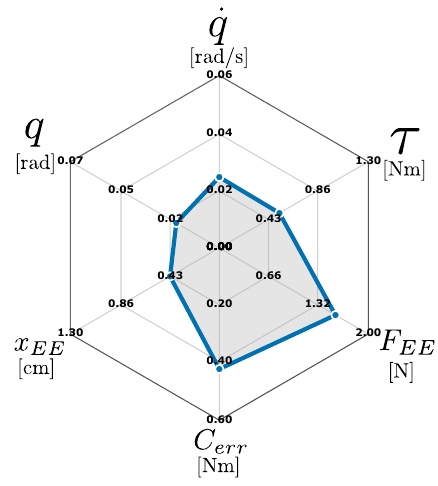}}\hfill
      \subfloat[Task 3]{\includegraphics[width=.48\columnwidth]{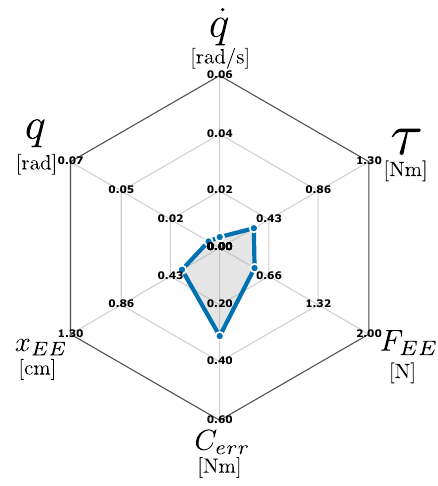}}\hfill
     \subfloat[Task 4]{\includegraphics[width=.48\columnwidth]{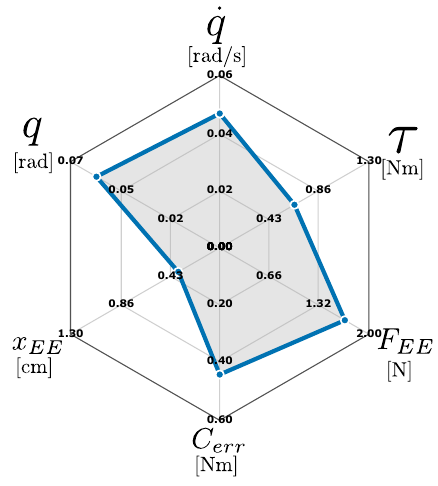}}\hfill
     \caption{Simulation error metrics results for the first four experiments: \\a) Free motion using a joint impedance example controller, b) Free motion using a Cartesian impedance example controller, c) Variable force holding, d) Circle trajectory following while maintaining constant contact force.}
    \label{fig:validation}     
\end{figure}

\begin{figure}[h]
       \subfloat[Left manipulator]{\includegraphics[width=.48\columnwidth]{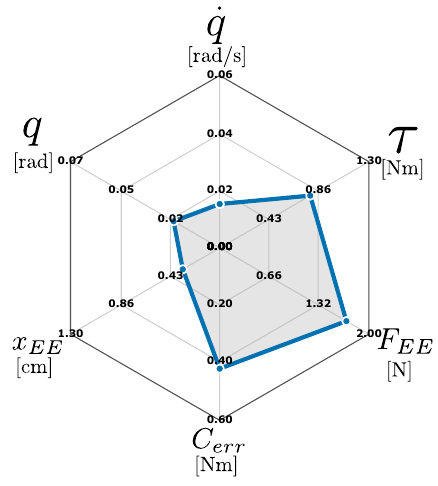}}\hfill
     \subfloat[Right manipulator]{\includegraphics[width=.48\columnwidth]{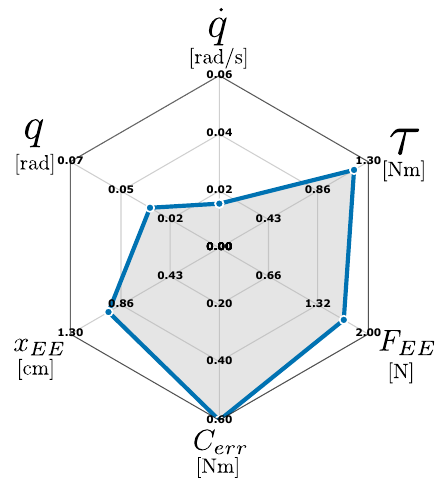}}\hfill
     \caption{Simulation error metrics results for the a) left and b) right manipulator during the bimanual box grasping task with constant contact force during lifting (task 5).}
    \label{fig:validation_exp5}
\end{figure}

The simulation fidelity metrics for the single-arm experiments are shown in Fig. \ref{fig:validation}, while results for the bimanual task are reported in Fig. \ref{fig:validation_exp5}. The results show that the discrepancy between simulation and reality is up to 0.06 rad, 0.05 rad/s, and 1.3 cm for joint positions, joint velocities, and EE position, respectively. For dynamics, the differences are up to 1.3 Nm for joint torques, 1.7 N for detected external forces, and 0.6 Nm for the control error. It is important to note that the real system exhibited noticeable sensor noise during no-contact tasks (tasks 1 and 2), which introduces a certain error for the $\tau$ and $F_{EE}$ metrics that is not considered by default in the simulation.
For the bimanual task (task 5), Fig. \ref{fig:validation_exp5} reveals pronounced discrepancies between the two manipulators, particularly in EE position, joint torques, and control error. Although both arms are of the same model, the differences indicate that simulation fidelity may vary substantially even among nominally identical systems. Likely causes include manipulator age and temperature-dependent effects on actuators and sensors. These observations underscore the necessity of incorporating real-system data to account for such variations when addressing the sim-to-real gap.

\begin{figure}[H]
\centering
    \includegraphics[width=0.55\columnwidth]{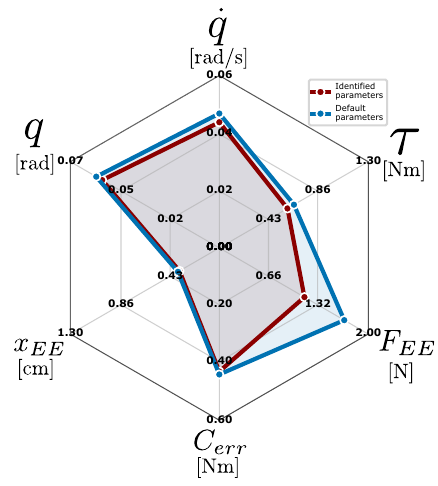}
     \caption{Simulation error metrics results comparison between the default (blue) and updated (red) simulation parameters during task 4 (circle trajectory following with constant contact force).}
    \label{fig:ident_comp}
\end{figure}

The final validation results, shown in Fig. \ref{fig:ident_comp}, illustrate the effect of updating simulation parameters through a real-world identification procedure before repeating experiment 4. The red line indicates a reduced error between the simulation and the real system compared to the default experiment (blue line). For instance, the recorded external force error is reduced by nearly half. These results complement the findings from task 5 and suggest that incorporating identification-based parameter tuning into the simulation can be an effective approach to narrowing the sim2real gap.
\newpage

\section{Discusion and Conclusion}
In this work, we introduced \texttt{multipanda\_ros2}, a novel architecture for real-time multi-robot control for the Franka Robotics robots within the ROS 2 framework. The framework integrates the full capabilities of the robot offered by \texttt{libfranka}, providing an easy entry to bimanual or multimanual torque control research to the community. 

First, we introduce an architecture that provides reliable, low-latency control, with controllet-feature switching delays of $\leq$2 ms. This capability is crucial for high-frequency applications and aligns with the stringent requirements of physical human-robot interaction and robot-environment modeling, which require 1 kHz command loops for safety and accurate dynamics \cite{Haddadin2022, ISO10218}. Our framework not only simplifies the implementation of complex controllers but also serves as a benchmark for reproducible research, mirroring the component-based designs.



Second, we integrate a high-fidelity MuJoCo simulation environment to tackle the sim2real gap. We provide a comprehensive set of metrics to evaluate both kinematic accuracy and dynamic consistency, a contribution that extends beyond the common practice of focusing on kinematic data alone \cite{2020ZhaoRL}. This approach allows researchers to assess simulation fidelity in terms of torques, forces, and control errors, which are critical for tasks involving physical contact.


Furthermore, we assess force-based fidelity by extending soft-robotics approaches \cite{Gao2024ResidualPhysics} to rigid dual-arm, contact-rich tasks. Through real-world inertial parameter identification, we demonstrate that incorporating data from physical systems can significantly reduce the sim2real gap, leading to improved force and torque accuracy in the simulated environment. This iterative physics refinement process offers a promising method for continuously improving simulation models, directly addressing a key limitation in current research, akin to recent approaches using residual physics networks \cite{Gao2024ResidualPhysics}.

While the proposed architecture demonstrates notable benefits, we acknowledge its limitations. Currently, the framework is exclusively tailored to the Franka platform and evaluated across a restricted set of validation scenarios. Furthermore, our sim2real refinement relies entirely on offline physical parameter identification, lacking the online adaptive capabilities necessary to compensate for dynamic, real-time environmental changes.

As future work, we plan to explore more complex validation scenarios and develop additional validation metrics that remain accessible to the broader robotics community. We will also investigate advanced techniques to further reduce the sim2real gap, thereby accelerating progress in robotics research. Specifically, we intend to leverage this tight integration to deploy the framework in digital twin applications, enabling real-time synchronization, monitoring, and predictive control of physical systems, e.g., \cite{Skerlj2024}. Finally, motivated by the strong interest already observed in the online repository, we will continue to expand and maintain this open-source framework to encourage wider adoption. Ultimately, we hope that \texttt{multipanda\_ros2} will be utilized to drive tangible, real-world use cases beyond the laboratory setting.
\newpage






%







\balance
\bibliographystyle{IEEEtran}
\bibliography{sections/tidy_bibliography}

\end{document}